\documentclass[conference]{IEEEtran}
\IEEEoverridecommandlockouts
\usepackage{cite}
\usepackage{amsmath,amssymb,amsfonts}
\usepackage{algorithmic}
\usepackage{graphicx}
\usepackage{textcomp}
\usepackage{xcolor}
\usepackage{url}
\usepackage{makecell}
\usepackage{balance}
\def\BibTeX{{\rm B\kern-.05em{\sc i\kern-.025em b}\kern-.08em
    T\kern-.1667em\lower.7ex\hbox{E}\kern-.125emX}}
\makeatletter
\newcommand{\linebreakand}{%
  \end{@IEEEauthorhalign}
  \hfill\mbox{}\par
  \mbox{}\hfill\begin{@IEEEauthorhalign}
}
\makeatother
\usepackage{caption}
\usepackage{subcaption}

\title{Sleep Stage Classification using Multimodal Embedding Fusion from EOG and PSM}

\author{\IEEEauthorblockN{Olivier Papillon}
\IEEEauthorblockA{\textit{Systems and Computer Engineering} \\
\textit{Carleton University}\\
olypapillon@cmail.carleton.ca}
\and
\IEEEauthorblockN{Rafik Goubran, Life Fellow IEEE}
\IEEEauthorblockA{\textit{Systems and Computer Engineering} \\
\textit{Carleton University}\\
\textit{Bruyère Health Research Institute}\\
goubran@sce.carleton.ca}
\and
\IEEEauthorblockN{James Green, Senior Member IEEE}
\IEEEauthorblockA{\textit{Systems and Computer Engineering} \\
\textit{Carleton University}\\
jrgreen@sce.carleton.ca}
\and
\IEEEauthorblockN{Julien Larivière-Chartier}
\IEEEauthorblockA{\textit{Bruyère Health Research Institute} \\
\textit{Systems and Computer Engineering} \\
\textit{Carleton University}\\
JLariviereChartier@bruyere.org}
\and
\IEEEauthorblockN{Caitlin Higginson}
\IEEEauthorblockA{\textit{Sleep Research Unit} \\
\textit{University of Ottawa Institute for Mental} \\
\textit{Health Research at the Royal Ottawa Hospital}\\
Caitlin.Higginson@theroyal.ca}
\and
\IEEEauthorblockN{Frank Knoefel, MD}
\IEEEauthorblockA{\textit{Systems and Computer Engineering} \\
\textit{Carleton University}\\
\textit{Bruyère Health Research Institute}\\
fknoefel@bruyere.org}
\linebreakand %
\IEEEauthorblockN{Rébecca Robillard}
\IEEEauthorblockA{\textit{School of Psychology,} \\
\textit{University of Ottawa} \\
\textit{University of Ottawa Institute for Mental} \\
\textit{Health Research at the Royal Ottawa Hospital}\\
Rebecca.Robillard@uottawa.ca}
}

\begin{document}
\maketitle

\begin{figure*}[tp]
    \includegraphics[width=\textwidth]{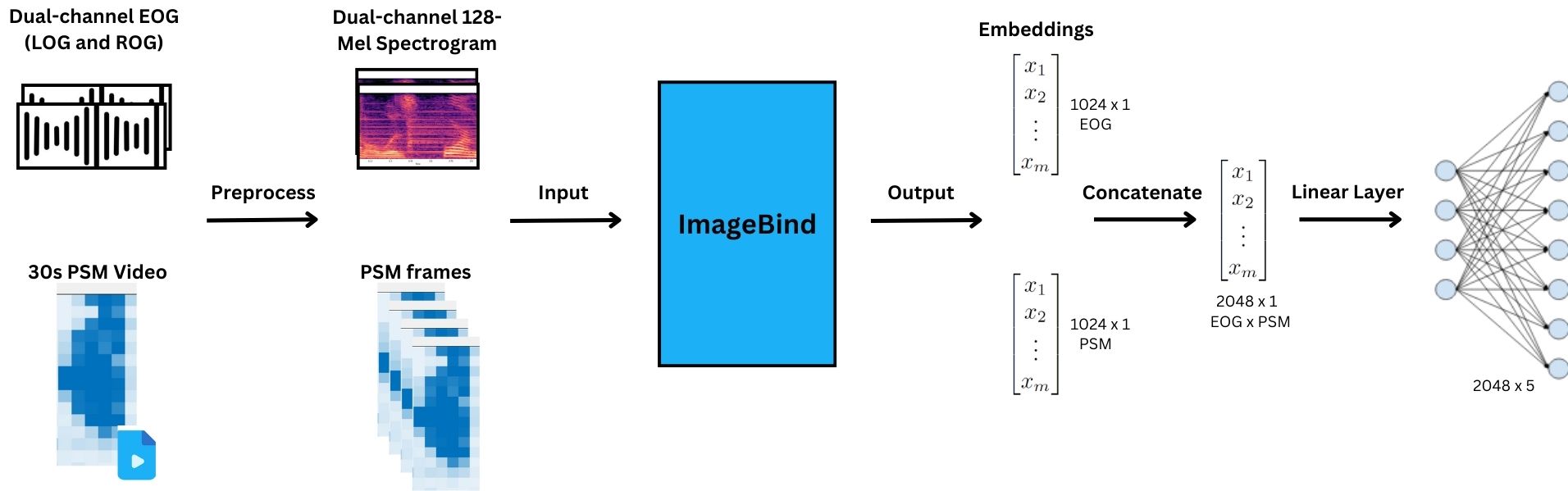}
\caption{Model Overview of the Multimodal Embedding Fusion Process for EOG and PSM}
\label{fig:ImageBind_Overview}
\end{figure*}
\begin{abstract}
Accurate sleep stage classification is essential for diagnosing sleep disorders, particularly in aging populations. While traditional polysomnography (PSG) relies on electroencephalography (EEG) as the gold standard, its complexity and need for specialized equipment make home-based sleep monitoring challenging. To address this limitation, we investigate the use of electrooculography (EOG) and pressure-sensitive mats (PSM) as less obtrusive alternatives for five-stage sleep-wake classification. This study introduces a novel approach that leverages ImageBind, a multimodal embedding deep learning model, to integrate PSM data with dual-channel EOG signals for sleep stage classification. Our method is the first reported approach that fuses PSM and EOG data for sleep stage classification with ImageBind. Our results demonstrate that fine-tuning ImageBind significantly improves classification accuracy, outperforming existing models based on single-channel EOG (DeepSleepNet), exclusively PSM data (ViViT), and other multimodal deep learning approaches (MBT). Notably, the model also achieved strong performance without fine-tuning, highlighting its adaptability to specific tasks with limited labeled data, making it particularly advantageous for medical applications. We evaluated our method using 85 nights of patient recordings from a sleep clinic. Our findings suggest that pre-trained multimodal embedding models, even those originally developed for non-medical domains, can be effectively adapted for sleep staging, with accuracies approaching systems that require complex EEG data.
\end{abstract}


\section{Introduction}

Sleep disorders are affecting a large number of Canadians, with important consequences for several other aspects of physical and mental health. However, there has been a long-standing issues with access to sleep tests, creating barriers for efficient sleep healthcare. Remote health monitoring could alleviate strain on the Canadian healthcare system by reducing the need for hospital beds, addressing long wait lists and high hospitalization costs.\cite{HMonitor}. Polysomnography (PSG), the study of sleep, becomes increasingly important as sleep-related disorders become more prevalent with age. Sleep stage classification is the process of identifying the different stages of sleep an individual is in. Research has shown that sleep disorders are exacerbated during specific stages of sleep; for example, sleep apnea is more common during Stage 1 Non-REM (NREM1), Stage 2 Non-REM (NREM2), and REM \cite{SleepCleveland}. Accurate sleep stage classification enables researchers to select the most appropriate estimation approaches for downstream tasks, such as automatic sleep apnea event detection. The state-of-the-art in automatic sleep stage classification uses single-channel electroencephalography (EEG) data from the differential signal between the electrode at FpZ (prefrontal midline location) and Cz (central midline location).\cite{AttentionEEGEldele} However, despite their high performance, the use of EEG-based sensors poses several practical challenges, including the need for trained clinicians, extensive skin preparation to reduce skin electrode electrical impedance, and high-level laboratory infrastructure to provide controlled environments. 
Due to the complexity of EEG sensors, they pose challenges for remote health monitoring. Therefore, we focus on less intrusive sensor modalities, such as pressure-sensitive mats (PSMs) and electrooculography (EOG). PSMs measure the contact pressure between the patient and the mattress over space and time. PSMs have been shown to be effective in multiple health monitoring applications, including sleep apnea detection \cite{PSM_Sleep_Apnea_Azimi}, vital sign estimation \cite{Bekele_Respiration,AzimiBreating}, patient movement detection \cite{foubert_posture_2010, FoubertLying,Bennett_Mobility}, false alarm detection \cite{Kyro_FalseALARM} and sleep stage classification \cite{UnobtrusiveSLEEP,SleepWakeClassDeep}. The EOG signal can be measured with an electrode placed on the skin near the lateral and medial canthus represents potential changes between the cornea and retina during eye movements. Accurate recording of eye movements is valuable because rolling eye movements signal the transition from wakefulness to light sleep, and rapid eye movements characterize REM sleep \cite{SMOLLEYEOG}. Moreover, it has been applied for fatigue tracking \cite{XuRealEye}, for sleep monitoring using eye tracking masks \cite{LiangEyeMask}, and for sleep stage classification \cite{EOGJiahao, RahmanEOG, MaitiEnhancing}. Researchers have proposed using less obtrusive sensor modalities and more easily deployable sensors in home environments for effective sleep monitoring of patients.\cite{HUANG_Unobstrusive} Among these modalities, EOG and PSM stand out as unobtrusive and more practical options for use in home-based sleep monitoring systems.

In the present study, we focused on automatically recognizing sleep stages from data recorded by polysomnography, using PSM and EOG data. To achieve this, we used machine learning (ML) to fuse and analyze both PSM and EOG data, extracting meaningful patterns and valuable insights. Recent research has determined sleep stages using EOG with deep learning architectures \cite{DeepSleepNet, EOGJiahao,MaitiEnhancing}. Although these methods have proven effective, recent deep learning architectures that integrate multiple modalities have demonstrated improved classification performance in machine learning tasks.\cite{li2024reviewdeeplearningbasedinformation} We hypothesize that the fusion of PSM data, which captures larger body movements, and EOG, which captures eye movements, will be more effective than either modality in isolation. Since PSM data represent images over time (i.e., video) and EOG signals are bioelectrical signals (i.e., analogous to audio signals), our data can be represented as a combination of video and audio signals, and therefore access deep learning models intended for fusing audio-video data \cite{ImageBind, NagraniAttention}.

Although deep learning methods are very powerful, training these models still poses a significant challenge due to the large amount of labelled data required \cite{ZhangDisease}. To overcome the current shortage of labelled training data, we utilized transfer learning to leverage state-of-the-art pre-trained deep learning models and fine-tune them for our specific task. In this paper, we evaluate the state-of-the-art in EOG-based sleep stage classification on data from actual patients; we demonstrate that ImageBind embeddings of EOG data outperform the state-of-the-art for sleep stage classification models; ultimately, we show that fusing EOG and PSM data using ImageBind multi-modal embeddings leads to further improvements in sleep stage classification accuracy relative to these models. This research contributes to the fields of computer vision, machine learning, and medical measurement and applications by examining the effectiveness of EOG and PSM data in detecting sleep stages. The combination of PSM and EOG requires minimal installation and respects user privacy. These characteristics make them well-suited for creating non-invasive sleep-tracking solutions for remote patient monitoring. 



\section{Related Works}
\subsection{Deep Learning using Pressure Sensor Mat Measurements}
In \cite{FoubertLying}, Foubert \textit{et al.} performed sleep position classification using classical machine learning (ML) methods. The researchers extracted features from the PSM data (e.g., the sum of sensor values and the center of pressure) to train SVM and K-NN classifiers for recognizing six different sleeping poses. In \cite{SleepWakeClassDeep}, Green \textit{et al.} also used PSM data for recognizing sleep versus wakefulness using a TCN network.

\subsection{Multi-Modal Sleep Stage Classification}
In \cite{AdvancingLin}, Lin \textit{et al.} developed a multimodal sleep stage classification method using EOG and EEG data from 100 participants across two datasets (SleepEDF-ST and SleepEDF-SC)\cite{Kemp_SleepEDF}. The study employed a hybrid deep learning architecture that combined multimodal local feature extraction, holistic correlation encoding, and cross-domain interaction modules to classify five stages of sleep-wake (Wake, N1, N2, N3, and REM).

\subsection{Sleep Stage Classification using EOG data}
In \cite{MaitiEnhancing}, Maiti \textit{et al.} proposed a novel SE-Resnet-Transformer model for sleep stage classification using single-channel EOG data. The study leveraged the EOG modality for automated classification of five stages of sleep-wake (Wake, N1, N2, N3, and REM). In \cite{EOGJiahao}, Jiahao \textit{et al.} introduced EOGNet, a deep learning model designed for sleep stage classification using EOG data. The model combines a two-scale Convolutional Neural Network (CNN) for feature extraction with a bidirectional Gated Recurrent Unit (Bi-GRU) to capture sequential information in sleep epochs. A two-step training protocol is employed: first, the CNN is pre-trained on balanced datasets, and then the entire model is fine-tuned with sequential inputs.

\subsection{Audio-Visual and Video Deep Learning}
In terms of audio-visual deep learning, in \cite{NagraniAttention}, Nagrani \textit{et al.} introduced a transformer-based model, Multimodal Bottleneck Transformer (MBT), to improve audio-visual fusion for video classification. While traditional models relied on late fusion, instead, MBT used a small set of latent units that form an attention bottleneck, which forces the model, within a given layer, to collate and condense information from each modality before sharing it with the other, reducing computational costs while improving performance. In \cite{ViVit}, Arnab \textit{et al.} introduced transformer-based models for video classification, replacing convolutional neural networks with self-attention to capture long-range dependencies. By leveraging pretraining on large image datasets, they trained ViViT on relatively small video datasets. It was evaluated on benchmarks such as Kinetics 400 \cite{kay2017kinetics}, where ViViT achieved state-of-the-art performance.



\subsection{ImageBind}
In \cite{ImageBind}, Girdhar \textit{et al.} introduced a model that learned a joint multimodal embedding space across six different modalities: images, text, audio, depth, thermal, and IMU data. Unlike prior approaches that required explicit multimodal pairs for training, ImageBind leverages only image-paired data by using the binding property of images to align all modalities into a shared representation. In the present study, we fine-tuned ImageBind to align EOG and PSM data and used it for sleep stage classification.

\subsection{DeepSleepNet}
In \cite{DeepSleepNet}, Supratak \textit{et al.} presented a deep learning-based model for automated sleep stage classification using raw single-channel EEG or EOG data. Unlike previous methods that relied on hand-engineered features, DeepSleepNet employed convolutional neural networks (CNNs) for time-invariant feature extraction and bidirectional long short-term memory (Bi-LSTM) networks to capture temporal dependencies, including sleep stage transitions. It is widely recognized as a state-of-the-art model in sleep stage classification. In the present study, this method served as a baseline for comparing the effectiveness of our multimodal sleep classification model.

\section{Methods}
\subsection{Clinical Data Collection}
During an overnight study conducted at the Sleep Disorders Clinic at The Royal Ottawa Hospital a combination of electroencephalography (EEG), electromyography (EMG), electrooculography (EOG), and bed-based pressure-sensing mats (PSM) that record contact pressure data throughout the night has been installed to identify different stages of sleep during sleep studies. Consenting participants were monitored in two rooms equipped with PSM as well as standard PSG sensors, including EEG, EOG, and EMG. Sleep data were continuously recorded and later annotated by a registered sleep technologist based on standard criteria \cite{AASM} with both sleep position (four poses) and sleep stage (five stages) information. The EOG sensors recorded the patients' eye movements, with one sensor for each eye (left and right). These sensors output values ranging from -3000 to 3000, which were then normalized to a scale of 0 to 1. Our dataset consists of 85 nights of patient data recordings. For the first 74 participants, data were sampled at 250 Hz;  for the remaining 11 participants, the sampling rate was raised to 512 Hz due to an equipment upgrade at the hospital. Each bed in the clinic is outfitted with several PSMs manufactured by Tactex Controls Inc. These mats consist of a 3 × 8 grid of fibre optic proximity sensors that detect pressure at a sampling rate of 10 Hz. The sensor values, which range from 0 to 2046, are normalized to a scale of 0 to 1. Each room has three PSMs connected to a data acquisition box (DAQ) for the upper body and three PSMs for the mid-to-lower body connected to another DAQ, resulting in an 18 × 8 low-resolution image output. The dataset comprises five sleep-wake stages: Wake, Non-REM Stage 1 (NREM1), Non-REM Stage 2 (NREM2), Non-REM Stage 3 (NREM3), and Rapid Eye Movement (REM).

\subsection{Preprocessing}
Our dataset comprised overnight sleep recordings for 85 subjects, with an average sleep duration of approximately 6.5 h per night. Sleep stage labels were provided for each 30 s epoch; therefore,  the PSM and EOG data were also split into non-overlapping 30 s epochs, under the assumption that the sleep stage remains constant over each epoch. This resulted in approximately 780 epochs per patient. With 85 research participants, our final dataset comprises 63,236 epochs. The PSM data were collected at 10 Hz. EOG signals were sampled at either 250 Hz or 512 Hz (differed by patient). For encoding by ImageBind, the EOG data were stored as dual-channel WAV audio files, with the left and right channels corresponding to the respective channels in the file. Thus, each record in the dataset corresponded to a 30 s epoch and comprised one PSM video and one dual-channel EOG wav file. 

For model evaluation, we adopted a five-fold cross-validation scheme to ensure robustness. Each fold comprises approximately 17 patients, resulting in a training set of 68 patients and a testing set of 17 patients per fold. Data from the same patient were never used for both training and testing, thereby avoiding overfitting

\subsection{Multimodal Embedding Fusion}
To fuse the EOG and PSM modalities, we aligned them using ImageBind embeddings. The ImageBind model uses a shared embedding space to bind multiple modalities—including images, audio, text, thermal, depth, and IMU signals, to an image-pair representation. First, each modality's data is preprocessed to be an acceptable input for ImageBind. For EOG, the WAV files are upsampled to 16 kHz and then converted into a 128-mel spectrogram. For PSM, each 30 s video is converted into individual frames for processing. These modalities are then fed into the pre-trained ImageBind model using their specified modality (i.e., audio for EOG and video for PSM). We evaluated two configurations: one with linear probing of ImageBind (i.e., the weights remain fixed) and another with ImageBind fine-tuned. The model outputs a vector (or embedding) of size 1024x1 for each modality. We then concatenate these two vectors to form a vector of size 2048x1 and add a linear layer to map the fused representation to a five class classification. This is illustrated in Figure \ref{fig:ImageBind_Overview}.

\begin{figure}[b]
     \centering
     \begin{subfigure}[b]{0.2\textwidth}
         \centering
         \includegraphics[width=\textwidth]{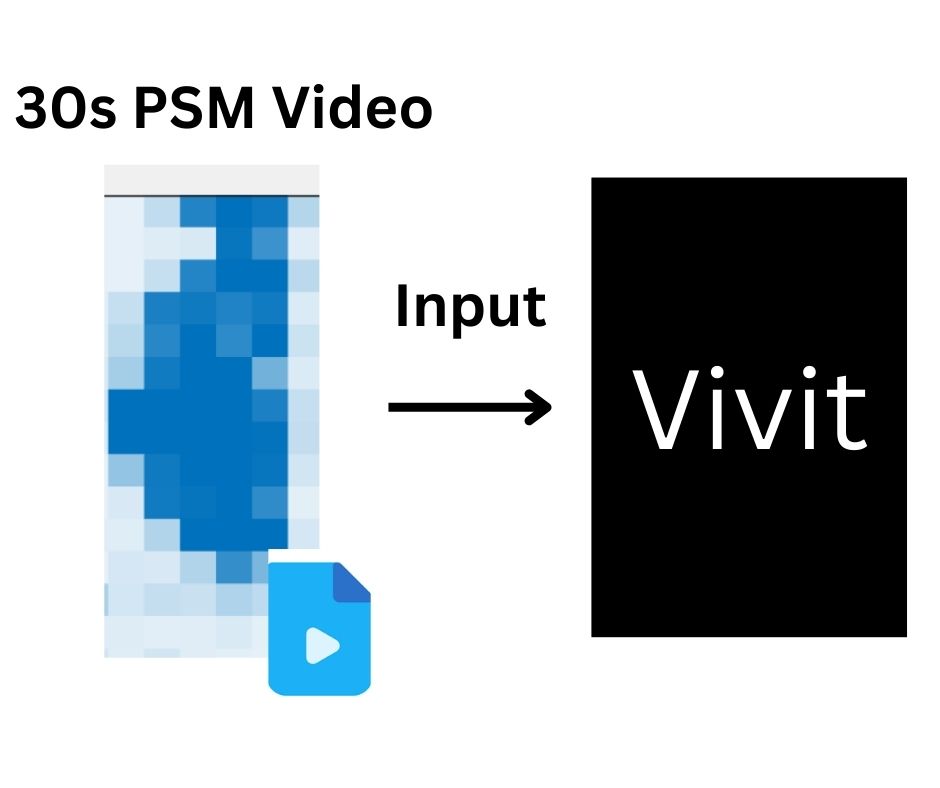}
         \caption{$ViVit$ }
         \label{fig:ViVit}
     \end{subfigure}
     \hfill
     \begin{subfigure}[b]{0.2\textwidth}
         \centering
         \includegraphics[width=\textwidth]{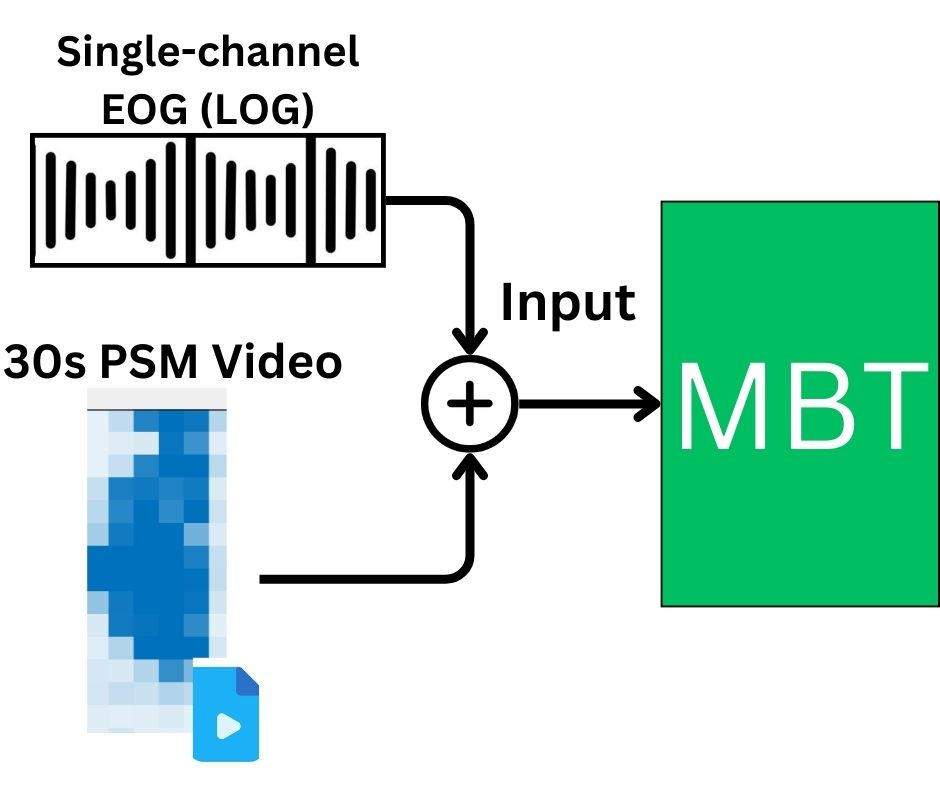}
         \caption{$MBT$ }
         \label{fig:MBT}
     \end{subfigure}
     \hfill
     \begin{subfigure}[b]{0.2\textwidth}
         \centering
         \includegraphics[width=\textwidth]{ 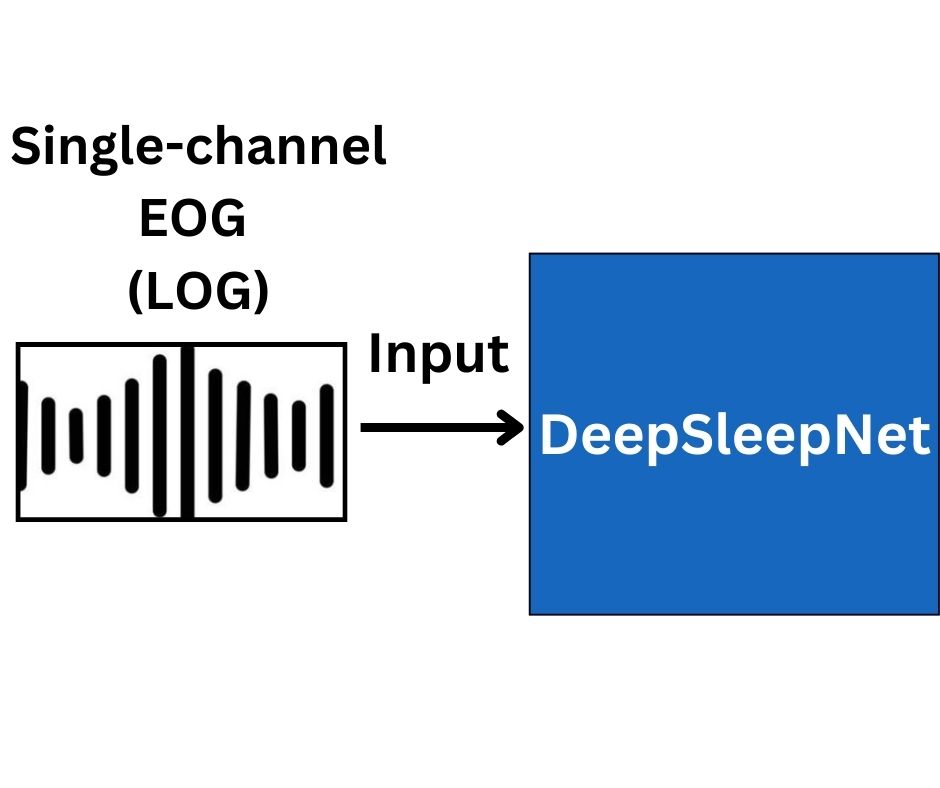}
         \caption{$DeepSleepNet$}
         \label{fig:DeepSleepNe}
     \end{subfigure}
      \hfill
     \begin{subfigure}[b]{0.2\textwidth}
         \centering
         \includegraphics[width=\textwidth]{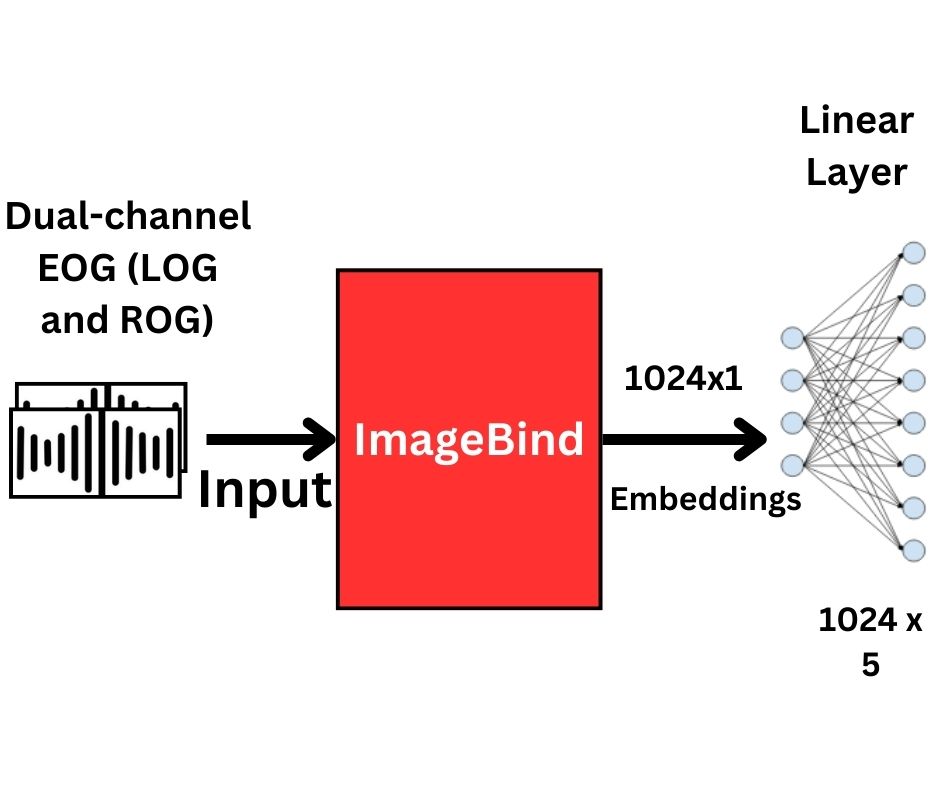}
         \caption{$ImageBind$ ($EOG$ $only$)}
         \label{fig:EOG_ONLY}
     \end{subfigure}
        \caption{Model overview of the input modalities for the three models: ViViT\cite{ViVit}, MBT\cite{NagraniAttention}, DeepSleepNet\cite{DeepSleepNet}, ImageBind with EOG only as an input}
        \label{fig:three graphs}
\end{figure}
\subsection{Machine Learning Methods}
We employed several pre-trained models, which operated on single or multiple modalities. 

As mentioned previously, we used two ImageBind models: one that is fine-tuned on both inputs and another where the weights are frozen (linear probing). In addition, we employed an ImageBind model that processes EOG inputs only. (Figure \ref{fig:EOG_ONLY}). In this configuration, the output vector (1024×1) is fed into a linear layer with dimensions 1024×5.

The ViViT model (Figure \ref{fig:ViVit}) is a state-of-the-art model pre-trained on the Kinetics-400 \cite{kay2017kinetics} dataset, which consists of 400 10 s RGB video clips covering 400 classes and having a resolution of 224×224 pixels. ViViT is designed to process video clips, therefore, we trained it using PSM video. However, the data produced by the PSM are normalized pressure readings with a single channel and a resolution of only 18×8 pixels. To make the data compatible with the model, we modified the transformer’s positional embeddings to support a smaller video input size and summed the three-channel RGB input weights into one channel to accommodate the single-channel PSM data.

The MBT model (Figure \ref{fig:MBT}) is a state-of-the-art multimodal model that employs two backbone models, one for video and one for audio. The video model is a ViT pre-trained on ImageNet-21k \cite{ridnik2021imagenet}, which comprises 14 million images across 21,843 classes where RGB images are 224×224 pixels. The audio model is also a ViT pre-trained on AudioSet\cite{AudioSet}, which consists of 632 audio classes derived from human-labeled 10 s sound clips, and it takes as input a 128-mel spectrogram of the audio clips. To make the data compatible with MBT, we converted the PSM videos to RGB and upscaled them using bilinear interpolation to 224×224 pixels, and we converted a single channel of the LOG (left EOG) into a 128-mel spectrogram.

The top-3 methods on the EOG-based sleep stage classification leaderboard are EOGNet\cite{EOGJiahao}, SeResNet\cite{MaitiEnhancing}, and DeepSleepNet\cite{DeepSleepNet}. Despite attempts to contact the authors of the first two methods, we were unable to reproduce their results. However, we have included DeepSleepNet in our performance comparison, since it represents the state of the art in EOG-based sleep stage classification. For DeepSleepNet, we fine-tuned the model to our clinical dataset. Since DeepSleepNet expects inputs of 3,000 samples and a single-chanel, we downsampled the LOG signal.

\subsection{Model Development}
All models were run on an A100 GPU using PyTorch. A hyperparameter sweep of the learning rate, weight decay, and number of epochs was performed to identify the configuration that yielded the highest average five-way accuracy across the five folds. All models were fine-tuned on our clinical dataset.

\section{Results and Discussion}
We compared different ML architectures for sleep stage classification using PSM only, single-channel EOG, and both modalities. Five-way classification accuracy and the macro-averaged F1 score were averaged over five folds. Although K-fold cross-validation ensures that the same data are never used for both training and testing a model, repeated cross-validation during hyperparameter optimization can sometimes lead to overfitting. The models developed here will be further validated on future patient data during continued data collection.

The results from Table \ref{tab:Results_Sleep} demonstrated that fine-tuning ImageBind and using PSM and dual-channel EOG outperformed all other methods. Notably, the model utilizing only PSM video performed worse than other modalities. However, incorporating the EOG sensor modality alone led to a substantial improvement in model performance. This improvement can be attributed to the strong correlation between sleep stages and EOG signals, particularly in distinguishing between REM and NREM stages. Furthermore, including multiple modalities, by combining PSM with the right EOG, provided additional features that further enhanced sleep stage classification. These findings suggest that multimodal approaches, such as ImageBind, can significantly improve sleep stage classification by effectively leveraging complementary information from different data modalities.
\begin{table}
\centering
\caption{Comparison of results for sleep stage classification on our clinical dataset}
\begin{tabular}{cccc}
        Method & Modality & Accuracy & F1 (macro)\\ \hline
         ViVit & PSM video &  0.399 & 0.164\\
         MBT & \makecell{PSM video \& \\ Single-channel EOG}  & 0.631& 0.543\\
         DeepSleepNet & \centering{Single-channel EOG} & 0.743& 0.682\\
         ImageBind & \makecell{PSM video \& \\ Dual-channel EOG}  & \textbf{0.745} & \textbf{0.683}\\ 
    \end{tabular}
    \label{tab:Results_Sleep}
\end{table}

The optimal hyperparameters for our best performing ImageBind using PSM and dual-channel EOG, classification model were as follows: we used a learning rate scheduler with an initial learning rate of 1.4e-7, weight decay of 0.005, and a batch size of 12. We used the AdamW optimizer with fine-tuning for 6 epochs.

\subsection{Comparaison of Training Methods for ImageBind}
We trained ImageBind using two different methods, as presented in Table \ref{tab:Results_ImageBind}. First, we linearly probed ImageBind to evaluate its domain adaptability, allowing us to assess how well it classifies sleep stages without fine-tuning. Notably, ImageBind performed well even without fine-tuning, demonstrating its ability to adapt to a specific task with limited labeled data, making it particularly suitable for medical applications. This is particularly impressive given that the model processes PSM videos, which are low-resolution pressure videos and EOG data as an audio input, which differs from its expected input modalities. Furthermore, we observed the incremental benefit of incorporating PSM video data, the accuracy improved from 0.710 to 0.745, while the F1-score increased from 0.636 to 0.683, highlighting the benefit of adding PSM data in enhancing sleep stage classification performance.

\begin{figure}[t]
     \centering
     \begin{subfigure}[b]{0.29\textwidth}
         \centering
         \includegraphics[width=\textwidth]{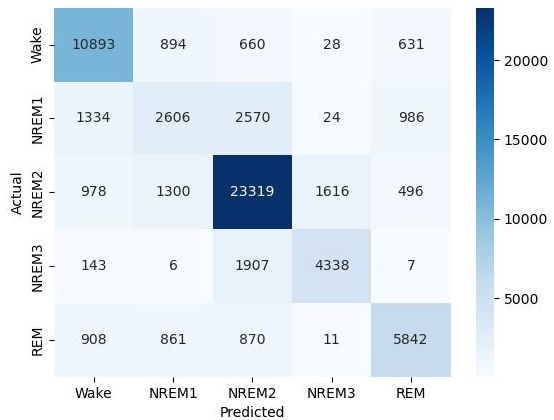}
         \caption{Confusion Matrix for ImageBind PSM and EOG}
         \label{fig:Imagebind-conf-FT}
     \end{subfigure}
     \hfill
     
     \begin{subfigure}[b]{0.29\textwidth}
         \centering
         \includegraphics[width=\textwidth]{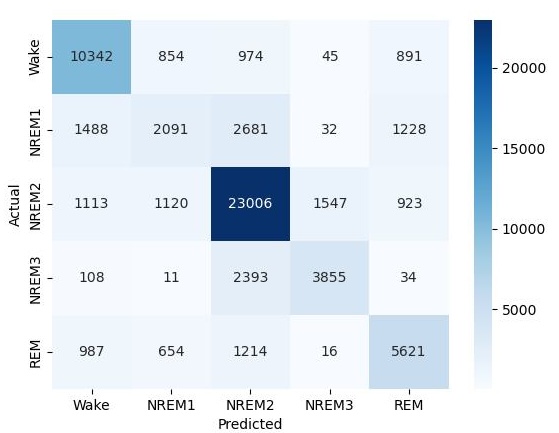}
         \caption{Confusion Matrix for ImageBind EOG-only}
         \label{fig:Imagebind-conf-EOG}
     \end{subfigure}
        \caption{The Confusion Matrix for ImageBind using Two Input Modalities: Fusion of PSM and EOG, and EOG-only}
        \label{fig:ImageBind-Conf}
\end{figure}

Comparing the confusion matrices from Figure \ref{fig:Imagebind-conf-FT} and Figure \ref{fig:Imagebind-conf-EOG}, we observed that both models performed well in distinguishing REM, Wake, and NREM2 stages. Incorporating PSM data further reduced classification errors, improving overall model accuracy. The primary challenge remained in differentiating NREM2, NREM3, and NREM1, which is inherently difficult when relying solely on EOG and PSM data. Additionally, most misclassifications occurred between adjacent sleep stages (e.g., NREM1 with Wake and NREM2, but not with NREM3 or REM). This is expected, as sleep stages transition gradually, making nearby stages more difficult to distinguish, whereas differentiating REM from Wake was relatively easier.

Our method is the first reported approach that uses PSM and EOG data for sleep stage classification with ImageBind. We demonstrated that employing pre-trained models from different modalities or domains improves sleep stage classification, even when there is a domain shift between the pre-training task (RGB video classification and sound classification) and the target domain (single-channel low-resolution PSM data and multi-channel EOG). This further illustrates the benefit of using state-of-the-art pre-trained models for medical applications, particularly when limited data are available.

\begin{table}
\centering
\caption{Comparison of results from ImageBind using different training methods for sleep stage classification for our clinical dataset}
\begin{tabular}{ccccc}
        Model & Method & Modality & Accuracy & F1 (macro)\\ \hline
        ImageBind & Fine-tuned & Dual-channel EOG  & 0.710 & 0.636 \\
        ImageBind &  
        Linear Probing & \makecell{PSM video \& \\ Dual-channel EOG}  & 0.690 & 0.614\\
    \end{tabular}
    \label{tab:Results_ImageBind}
\end{table}

\section{Conclusion}
Our study demonstrated that using multimodal deep learning models can enhance sleep stage classification from PSM and multi-channel EOG data through fine-tuning using data collected from actual patients in a clinical overnight sleep study. We showed that incorporating PSM improves the classification of sleep stages in individuals with sleep complaints. Additionally, the model exhibited strong domain adaptability, effectively learning from limited labeled data, which is particularly advantageous for medical applications where annotated datasets are often scarce. These results suggest that a fine-tuned ImageBind model holds significant potential for practical sleep stage estimation in home environments.

In future work, we will explore the integration of position estimation and sleep stage classification for applications, such as sleep apnea detection or vital sign estimation, and validate the results on high-resolution PSM data.

\section*{Acknowledgment}
We acknowledge the support of the Natural Sciences and Engineering Research Council of Canada (NSERC).

F. Knoefel acknowledges the support for the Bruyère Health Research Institute Chair in Research in Technology for Aging in Place.

This study followed IRB-approved protocol  \#112123 ``Sleeping and Health Improvements; Following Longitudinal Trajectories (SHIFT)'' 
\nocite{PhysioNet}
\balance
\bibliographystyle{ieeetr} 
\bibliography{sources}    

\begin{thebibliography}{10}

\bibitem{HMonitor}
M.~Cohen-McFarlane, S.~Bennett, B.~Wallace, R.~Goubran, and F.~Knoefel, ``{Bed-Based Health Monitoring Using Pressure Sensitive Technology: A Review},'' {\em IEEE Instrumentation \& Measurement Magazine}, vol.~24, no.~2, pp.~13--23, 2021.

\bibitem{SleepCleveland}
C.~Clinic, ``{Sleep Basics},'' 2024.
\newblock \url{https://my.clevelandclinic.org/12148-sleep-basics}.

\bibitem{AttentionEEGEldele}
E.~Eldele, Z.~Chen, C.~Liu, M.~Wu, C.-K. Kwoh, X.~Li, and C.~Guan, ``{An Attention-Based Deep Learning Approach for Sleep Stage Classification With Single-Channel EEG},'' {\em IEEE Transactions on Neural Systems and Rehabilitation Engineering}, vol.~29, pp.~809--818, 2021.

\bibitem{PSM_Sleep_Apnea_Azimi}
H.~Azimi, P.~Xi, M.~Bouchard, R.~Goubran, and F.~Knoefel, ``{Machine Learning-Based Automatic Detection of Central Sleep Apnea Events From a Pressure Sensitive Mat},'' {\em IEEE Access}, vol.~8, pp.~173428--173439, 2020.

\bibitem{Bekele_Respiration}
A.~Bekele, S.~Nizami, Y.~S. Dosso, C.~Aubertin, K.~Greenwood, J.~Harrold, and J.~R. Green, ``{Real-time Neonatal Respiratory Rate Estimation using a Pressure-Sensitive Mat},'' in {\em 2018 IEEE International Symposium on Medical Measurements and Applications (MeMeA)}, pp.~1--5, 2018.

\bibitem{AzimiBreating}
H.~Azimi, S.~Soleimani~Gilakjani, M.~Bouchard, S.~Bennett, R.~A. Goubran, and F.~Knoefel, ``{Breathing signal combining for respiration rate estimation in smart beds},'' in {\em 2017 IEEE International Symposium on Medical Measurements and Applications (MeMeA)}, pp.~303--307, 2017.

\bibitem{foubert_posture_2010}
N.~Foubert, {\em {Posture recognition and postural transition detection using bed-based pressure sensor arrays}}.
\newblock Master of {Applied} {Science}, Carleton University, Ottawa, Ontario, 2010.

\bibitem{FoubertLying}
N.~Foubert, A.~M. McKee, R.~A. Goubran, and F.~Knoefel, ``{Lying and sitting posture recognition and transition detection using a pressure sensor array},'' in {\em 2012 IEEE International Symposium on Medical Measurements and Applications Proceedings}, pp.~1--6, 2012.

\bibitem{Bennett_Mobility}
S.~Bennett, Z.~Ren, R.~Goubran, K.~Rockwood, and F.~Knoefel, ``{In-Bed Mobility Monitoring Using Pressure Sensors},'' {\em IEEE Transactions on Instrumentation and Measurement}, vol.~64, no.~8, pp.~2110--2120, 2015.

\bibitem{Kyro_FalseALARM}
D.~G. Kyrollos, K.~Greenwood, J.~Harrold, and J.~R. Green, ``{Detection of False Alarms in the {NICU} Using Pressure Sensitive Mat},'' in {\em 2021 IEEE Sensors Applications Symposium (SAS)}, pp.~1--5, 2021.

\bibitem{UnobtrusiveSLEEP}
L.~Samy, M.-C. Huang, J.~J. Liu, W.~Xu, and M.~Sarrafzadeh, ``{Unobtrusive Sleep Stage Identification Using a Pressure-Sensitive Bed Sheet},'' {\em IEEE Sensors Journal}, vol.~14, no.~7, pp.~2092--2101, 2014.

\bibitem{SleepWakeClassDeep}
C.~Green, M.~Bouchard, R.~Goubran, R.~Robillard, C.~Higginson, E.~Lee, and F.~Knoefel, ``{Sleep-Wake and Body Position Classification with Deep Learning using Pressure Sensor Mat Measurements},'' in {\em 2023 IEEE International Symposium on Medical Measurements and Applications (MeMeA)}, pp.~1--6, 2023.

\bibitem{SMOLLEYEOG}
L.~Smolley, ``{Adult polysomnography},'' in {\em Encyclopedia of Sleep and Circadian Rhythms (Second Edition)} (C.~A. Kushida, ed.), pp.~474--477, Oxford: Academic Press, second edition~ed., 2023.

\bibitem{XuRealEye}
J.~Xu, J.~Min, and J.~Hu, ``{Real-time eye tracking for the assessment of driver fatigue},'' {\em Healthcare Technology Letters}, vol.~5, pp.~54--58, Jan. 2018.

\bibitem{LiangEyeMask}
S.-F. Liang, C.-E. Kuo, Y.-C. Lee, W.-C. Lin, Y.-C. Liu, P.-Y. Chen, F.-Y. Cherng, and F.~Shaw, ``{Development of an EOG-Based Automatic Sleep-Monitoring Eye Mask},'' {\em IEEE Transactions on Instrumentation and Measurement}, vol.~64, pp.~1--1, 11 2015.

\bibitem{EOGJiahao}
J.~Fan, C.~Sun, M.~Long, C.~Chen, and W.~Chen, ``{EOGNET: A Novel Deep Learning Model for Sleep Stage Classification Based on Single-Channel EOG Signal},'' {\em Frontiers in Neuroscience}, vol.~15, 2021.

\bibitem{RahmanEOG}
M.~M. Rahman, M.~I.~H. Bhuiyan, and A.~R. Hassan, ``{Sleep stage classification using Single-Channel EOG},'' {\em Computers in Biology and Medicine}, vol.~102, pp.~211--220, 2018.

\bibitem{MaitiEnhancing}
S.~Maiti, S.~K. Sharma, and R.~S. Bapi, ``{Enhancing Healthcare with EOG: A Novel Approach to Sleep Stage Classification},'' in {\em 2024 IEEE International Conference on Acoustics, Speech and Signal Processing (ICASSP)}, pp.~2305--2309, 2024.

\bibitem{HUANG_Unobstrusive}
Z.~Huang, J.~Li, and Z.~He, ``{Full-coverage unobtrusive health monitoring of elders at homes},'' {\em Internet of Things}, vol.~26, p.~101182, 2024.

\bibitem{DeepSleepNet}
A.~Supratak, H.~Dong, C.~Wu, and Y.~Guo, ``{DeepSleepNet: A Model for Automatic Sleep Stage Scoring Based on Raw Single-Channel EEG},'' {\em IEEE Transactions on Neural Systems and Rehabilitation Engineering}, vol.~25, p.~1998–2008, Nov. 2017.

\bibitem{li2024reviewdeeplearningbasedinformation}
Y.~Li, M.~E.~H. Daho, P.-H. Conze, R.~Zeghlache, H.~L. Boité, R.~Tadayoni, B.~Cochener, M.~Lamard, and G.~Quellec, ``{A review of deep learning-based information fusion techniques for multimodal medical image classification},'' 2024.

\bibitem{ImageBind}
R.~Girdhar, A.~El-Nouby, Z.~Liu, M.~Singh, K.~V. Alwala, A.~Joulin, and I.~Misra, ``{Imagebind: One Embedding Space To Bind Them All},'' in {\em Proceedings of the IEEE/CVF Conference on Computer Vision and Pattern Recognition}, pp.~15180--15190, 2023.

\bibitem{NagraniAttention}
A.~Nagrani, S.~Yang, A.~Arnab, A.~Jansen, C.~Schmid, and C.~Sun, ``{Attention Bottlenecks for Multimodal Fusion},'' in {\em Advances in Neural Information Processing Systems} (M.~Ranzato, A.~Beygelzimer, Y.~Dauphin, P.~Liang, and J.~W. Vaughan, eds.), vol.~34, pp.~14200--14213, Curran Associates, Inc., 2021.

\bibitem{ZhangDisease}
Y.~Zhang, C.~Li, L.~Zhaoxia, and M.~Li, ``{Semi-Supervised Disease Classification Based on Limited Medical Image Data},'' {\em IEEE Journal of Biomedical and Health Informatics}, vol.~PP, pp.~1--12, 01 2024.

\bibitem{AdvancingLin}
Y.~Lin, Z.~Chen, L.~Ruan, H.~Luo, A.~Pumir, and J.~Xu, ``{Advancing Automated Sleep Stage Identification Through Multimodal PSG-Based Multiview Analysis},'' in {\em 2024 IEEE International Symposium on Medical Measurements and Applications (MeMeA)}, pp.~1--6, 2024.

\bibitem{Kemp_SleepEDF}
B.~Kemp, A.~Zwinderman, B.~Tuk, H.~Kamphuisen, and J.~Oberye, ``{Analysis of a sleep-dependent neuronal feedback loop: the slow-wave microcontinuity of the EEG},'' {\em IEEE Transactions on Biomedical Engineering}, vol.~47, no.~9, pp.~1185--1194, 2000.

\bibitem{ViVit}
A.~Arnab, M.~Dehghani, G.~Heigold, C.~Sun, M.~Lu\v{c}i\'c, and C.~Schmid, ``{ViViT: A Video Vision Transformer},'' in {\em Proceedings of the IEEE/CVF International Conference on Computer Vision (ICCV)}, pp.~6836--6846, October 2021.

\bibitem{kay2017kinetics}
W.~Kay, J.~Carreira, K.~Simonyan, B.~Zhang, C.~Hillier, S.~Vijayanarasimhan, F.~Viola, T.~Green, T.~Back, P.~Natsev, {\em et~al.}, ``{The Kinetics Human Action Video Dataset},'' {\em arXiv preprint arXiv:1705.06950}, 2017.

\bibitem{AASM}
R.~Berry, R.~Brooks, S.~Harding, R.~Lloyd, S.~Quan, M.~Troester, and B.~Vaughn, ``Aasm scoring manual updates for 2017 (version 2.4),'' {\em Journal of clinical sleep medicine : JCSM : official publication of the American Academy of Sleep Medicine}, vol.~13, 04 2017.

\bibitem{ridnik2021imagenet}
T.~Ridnik, E.~Ben-Baruch, A.~Noy, and L.~Zelnik-Manor, ``{Imagenet-21k pretraining for the masses},''

\bibitem{AudioSet}
J.~F. Gemmeke, D.~P.~W. Ellis, D.~Freedman, A.~Jansen, W.~Lawrence, R.~C. Moore, M.~Plakal, and M.~Ritter, ``{Audio Set: An ontology and human-labeled dataset for audio events},'' in {\em Proc. IEEE ICASSP 2017}, (New Orleans, LA), 2017.

\bibitem{PhysioNet}
A.~L. Goldberger, L.~A.~N. Amaral, L.~Glass, J.~M. Hausdorff, P.~C. Ivanov, R.~G. Mark, J.~E. Mietus, G.~B. Moody, C.-K. Peng, and H.~E. Stanley, ``{PhysioBank, PhysioToolkit, and PhysioNet: Components of a New Research Resource for Complex Physiologic Signals},'' {\em Circulation}, vol.~101, no.~23, pp.~e215--e220, 2000 (June 13).

\end{thebibliography}

\end{document}